\PassOptionsToPackage{unicode}{hyperref}
\PassOptionsToPackage{hyphens}{url}
\documentclass[
  11pt,
]{article}
\usepackage{xcolor}
\usepackage[margin=1in]{geometry}
\usepackage{amsmath,amssymb}
\usepackage[T1]{fontenc}
\usepackage[utf8]{inputenc}
\usepackage{textcomp} 
\usepackage[nott]{newtxtext}
\IfFileExists{upquote.sty}{\usepackage{upquote}}{}
\IfFileExists{microtype.sty}{
  \usepackage[]{microtype}
  \UseMicrotypeSet[protrusion]{basicmath} 
}{}
\makeatletter
\@ifundefined{KOMAClassName}{
  \IfFileExists{parskip.sty}{%
    \usepackage{parskip}
  }{
    \setlength{\parindent}{0pt}
    \setlength{\parskip}{6pt plus 2pt minus 1pt}}
}{
  \KOMAoptions{parskip=half}}
\makeatother
\usepackage{longtable,booktabs,array}
\usepackage{calc} 
\usepackage{etoolbox}
\makeatletter
\patchcmd\longtable{\par}{\if@noskipsec\mbox{}\fi\par}{}{}
\makeatother
\IfFileExists{footnotehyper.sty}{\usepackage{footnotehyper}}{\usepackage{footnote}}
\makesavenoteenv{longtable}
\usepackage{graphicx}
\makeatletter
\newsavebox\pandoc@box
\newcommand*\pandocbounded[1]{
  \sbox\pandoc@box{#1}%
  \Gscale@div\@tempa{\textheight}{\dimexpr\ht\pandoc@box+\dp\pandoc@box\relax}%
  \Gscale@div\@tempb{\linewidth}{\wd\pandoc@box}%
  \ifdim\@tempb\p@<\@tempa\p@\let\@tempa\@tempb\fi
  \ifdim\@tempa\p@<\p@\scalebox{\@tempa}{\usebox\pandoc@box}%
  \else\usebox{\pandoc@box}%
  \fi%
}
\def\fps@figure{htbp}
\makeatother
\providecommand{\tightlist}{%
  \setlength{\itemsep}{0pt}\setlength{\parskip}{0pt}}
\usepackage{lmodern}\usepackage{newtxmath}
\usepackage{bookmark}
\IfFileExists{xurl.sty}{\usepackage{xurl}}{} 
\makeatletter
\@ifundefined{xmpquote}{}{}
\makeatother
\hypersetup{
  pdftitle={Shared Misses, Individual Extras: Multi-Model Agreement and Document Length in LLM-Based Policy Evidence Analysis},
  pdfauthor={Ng S. T. Chong},
  hidelinks,
  pdfcreator={LaTeX via pandoc}}

\author{}
\date{}

\begin{document}

\begin{center}
{\LARGE\bfseries Shared Misses, Individual Extras: Multi-Model Agreement and Document Length in LLM-Based Policy Evidence Analysis\par}
\vspace{1.1em}
{\large Ng S. T. Chong\par}
\vspace{0.25em}
Campus Computing Centre, United Nations University, Tokyo, Japan\par
\vspace{0.15em}
\texttt{ngstc@unu.edu}\par
\end{center}
\vspace{0.8em}

\subsection{Abstract}\label{abstract}

Language models can check a policy document against a governance
framework one requirement at a time. We ask whether to run one model or
several, and whether long documents must be read in pieces. In a
small-scale study, four models assessed China's Personal Information
Protection Law (15 pages) and Malaysia's Personal Data Protection Act
(100 pages) against the 50 core elements of the Health Data Governance
Principles in six areas of law, 300 cells each. Two human coders, who
were not told what any model had concluded, checked 60 cells per law,
weighted back to all cells.

The models' misses were shared and their extra findings were individual.
Had the four missed evidence independently, together they would have
missed about 1\% of the evidence in China's law; they missed 22\%,
against 23\% for the most thorough model alone. Cells that only one
model flagged held evidence 4 times in 26; cells all four flagged held
it 18 times in 19. More models thus added little evidence to a thorough
model, but a strong signal of which findings to trust; added to a more
cautious model, the others raised recall by 10 to 27 points. Two runs of
the same model disagreed on about 30\% of the cells either flagged,
against about 45\% for two different models.

On the longer law no loss of recall was visible and wrong findings did
not become more frequent, because each cell asks for one qualifying
provision, not a complete list; misses and errors concerned which
element or area of law a passage serves, not unread pages. The study
supports using several models as a cross-check, not as a substitute for
judgment: their greatest value is helping people decide what to trust
provisionally, what to question, and where to look more closely.

\textbf{Keywords:} large language models; policy analysis; multi-model
ensembles; human review; document length; health data governance

\subsection{1. Introduction}\label{introduction}

Assessing whether a country's laws meet a governance framework is slow
work. For each requirement of the framework, and for each body of law
that might hold it, an analyst decides whether the country's legal texts
provide for it and cites the provision. For the Health Data Governance
Principles (Transform Health n.d.), whose assessment tool asks about 50
core elements across six areas of law, that is 300 judgments per
country, each needing a passage found and read. A language model can
produce a first pass in minutes. Two design questions then decide how
much that first pass is worth.

The first is whether to run one model or several. Running four models
costs about four times as much as running one. It is worth it only if
the four together find evidence that one would miss, or if their
agreement tells a reviewer something one model cannot.

The second is whether a long document must be read in pieces. An earlier
study of extracting adaptation actions from national climate plans found
that models reading a whole plan in one pass stopped short: every such
run cited at most 51\% of the pages it was given, and reading the same
plans in smaller pieces lifted recall to at least 0.918 in 77 of 81 runs
(Chong 2026b). If the same holds when checking laws against a framework,
long laws should be read in pieces too.

The two tasks ask different questions, though. Extraction asks the model
to list every action in a document; a long document gives it room to
stop listing early. The evidence task asks, for one element in one area
of law, whether there is at least one qualifying provision; a single
passage settles it. This paper tests how far the earlier findings carry
over.

We ran four models over two data protection laws of very different
length: China's Personal Information Protection Law (PIPL, 15 pages) and
Malaysia's Personal Data Protection Act 2010 (PDPA, 100 pages). Two
human coders, who did not see what any model had concluded, checked a
sample of 60 cells per law, drawn so that the cells where the models
disagreed were well covered. Each checked cell was weighted by how many
cells it stands for, giving estimates over all 300 cells of each law. A
second, identical run of each model, made for a separate experiment on
prompt rules, lets us compare the variation between models with the
variation between runs of one model.

This is a small-scale study: two laws, four models and 120 hand-checked
cells. Its scale limits how precisely it can measure and how far its
numbers generalize, not how soundly they were measured. Two coders
worked independently, without knowing what any model had concluded, and
agreed on 110 of the 120 cells; each of the other 10 was settled by a
written rule. The sample concentrates on the cells that decide the
questions asked, where the models disagree or none flags anything, and
every estimate is weighted back to all 300 cells of each law, with an
interval that shows its uncertainty. The main contrasts are large
relative to that uncertainty. Had the models' misses been independent,
about 1 of the PIPL's 79 evidence cells would have been missed by all
four; the estimate is 17, and at least 5 within the 95\% interval. Cells
all four models flagged held evidence 18 times in 19, against 4 times in
26 for cells one model flagged, and that pattern held on each law
separately. The findings describe these two laws and these four models,
not rates for any law or model: they show what to look for, and larger
studies are needed to measure it precisely.

The findings fit one pattern: the models' misses were shared, and their
extra findings were individual.

\begin{itemize}
\tightlist
\item
  \textbf{Several models found little evidence that the most thorough
  one missed.} Its recall was 1.3 points below the four models' combined
  recall on the PIPL and 3.4 points below on the PDPA. Had the models
  missed evidence independently, combining them would have recovered
  most of what each missed. They missed the same provisions.
\item
  \textbf{Their agreement is a strong signal of which findings to
  trust.} Cells all four models flagged held evidence 18 times in 19;
  cells one model flagged, 4 times in 26.
\item
  \textbf{What more models add depends on the base.} Added to a cautious
  model, the other three raised recall by 10 to 27 points; added to the
  most thorough one, by 1 to 3.
\item
  \textbf{Part of what looks like model diversity is run-to-run
  variation.} Two runs of the same model disagreed on about 30\% of the
  cells either flagged; two different models, on about 45\%.
\item
  \textbf{Length did not visibly cost recall, and the rate of wrong
  findings did not rise.} The pages no model cited in the long law were
  mostly its front matter and administrative machinery. The evidence
  models missed was on pages they had read.
\item
  \textbf{The dominant error is attribution:} crediting a passage to the
  wrong element or the wrong area of law. Writing the two rules that
  settled the coders' disagreements into the prompt helped the model
  that reported too much, and brought no reliable gain for the three
  that were already precise.
\end{itemize}

Taken together, the findings support using several models as a
cross-check, not as a substitute for judgment. Their greatest value is
helping people decide what to trust provisionally, what to question, and
where to look more closely.

Section 2 places the study in related work. Section 3 describes the task
and the tool, and Section 4 the data and the human coding. Sections 5 to
8 report what several models add, how runs compare with models, what
length does, and the test of written attribution rules. Section 9 turns
the findings into a workflow, and Section 10 states what the study
cannot show.

\subsection{2. Related work}\label{related-work}

\textbf{Language models as annotators.} Language models now annotate
text at a quality that rivals crowd workers on several social science
tasks (Gilardi et al.~2023), which makes them candidates for the first
pass of qualitative policy analysis. Work on model-assisted annotation
of subjective categories asks how far a model's suggestion shapes the
label a person gives, and when the combination is good enough (Schroeder
et al.~2025). We measure agreement between human coders, and between
models, with Krippendorff's alpha (Krippendorff 2004; Artstein and
Poesio 2008).

\textbf{Several models, or several samples of one.} A panel of smaller
models from different families can judge model outputs better and more
cheaply than one large judge (Verga et al.~2024), and majority votes
over model ensembles have been reported to beat individual models in
content categorization (Kamen and Kamen 2025) and cell-type annotation
(Yang et al.~2025). Self-consistency takes a majority over several
samples from one model (Wang et al.~2023). Our earlier work used
multi-model consensus for extracting and scoring climate policy (Chong
2026a) and found that councils of models corroborated rather than
extended the best single model's recall (Chong 2026b). This paper adds
human-coded evidence on why: whose misses overlap and whose findings do
not, and how much of the disagreement between models is variation
between runs.

\textbf{Long documents and policy text.} Models use information in the
middle of a long context less reliably than at its ends (Liu et
al.~2024), and retrieval tests that hide one fact in a long text became
a standard probe (Kamradt 2023). Information extraction from policy
documents with language models spans fiscal documents (Aggarwal et
al.~2025), food policy (Zhang et al.~2026) and wider surveys (Xu et
al.~2024). For list extraction from climate plans, whole-document
reading stopped short and reading in pieces fixed it (Chong 2026b).
Whether that carries over to a task that asks for one qualifying
provision has not been tested.

\subsection{3. The task and the tool}\label{the-task-and-the-tool}

\subsubsection{3.1 Cells and verdicts}\label{cells-and-verdicts}

The framework is the Health Data Governance Principles (Transform Health
n.d.). Its 8 principles, grouped under 3 objectives, hold 50 core
elements between them (Table 1). Each element names something a
country's rules should provide for, such as \emph{1.1.3 Collect personal
or sensitive data only when necessary and with informed consent} or
\emph{2.2.1 Establish data sharing rules and guidelines}.

\begin{longtable}[]{@{}
  >{\raggedright\arraybackslash}p{(\linewidth - 4\tabcolsep) * \real{0.3671}}
  >{\raggedleft\arraybackslash}p{(\linewidth - 4\tabcolsep) * \real{0.1139}}
  >{\raggedright\arraybackslash}p{(\linewidth - 4\tabcolsep) * \real{0.5190}}@{}}
\caption{\textbf{The framework's 8 principles and their core
elements.}}\tabularnewline
\toprule\noalign{}
\begin{minipage}[b]{\linewidth}\raggedright
Principle
\end{minipage} & \begin{minipage}[b]{\linewidth}\raggedleft
Core elements
\end{minipage} & \begin{minipage}[b]{\linewidth}\raggedright
One of them
\end{minipage} \\
\midrule\noalign{}
\endfirsthead
\toprule\noalign{}
\begin{minipage}[b]{\linewidth}\raggedright
Principle
\end{minipage} & \begin{minipage}[b]{\linewidth}\raggedleft
Core elements
\end{minipage} & \begin{minipage}[b]{\linewidth}\raggedright
One of them
\end{minipage} \\
\midrule\noalign{}
\endhead
\bottomrule\noalign{}
\endlastfoot
1.1 Protecting Individuals and Communities & 8 & 1.1.3 Collect personal
or sensitive data only when necessary and with informed consent \\
1.2 Building Trust in Data Systems & 8 & 1.2.5 Ensure data quality,
availability, and accessibility \\
1.3 Ensuring Data Security & 4 & 1.3.3 Ensure transparency around data
breaches \\
2.1 Enhance Health Systems and Services & 6 & 2.1.6 Enable and empower
frontline health workers \\
2.2 Promote Data Sharing and Interoperability & 7 & 2.2.2 Validate
informed consent before sharing data \\
2.3 Facilitate Innovation Using Health Data & 4 & 2.3.1 Apply health
data governance to emerging technologies \\
3.1 Promote Equitable Benefits From Health Data & 6 & 3.1.5 Implement
inclusive data feedback mechanisms \\
3.2 Establish Data Rights and Ownership & 7 & 3.2.2 Define clear
governance roles and responsibilities \\
\end{longtable}

Each element is assessed in six areas of law, which the tool calls
priority areas. Each area comes with a short description, which the
models see too:

\begin{itemize}
\tightlist
\item
  \textbf{Data protection and privacy:} rules on collecting, using,
  storing and sharing personal data, and individuals' rights over it,
  such as consent, access, correction and erasure.
\item
  \textbf{Cybercrime and cybersecurity:} security requirements, breach
  notification, incident response, and criminal provisions on
  unauthorized access to data or its misuse.
\item
  \textbf{General health law:} health system law that is not specific to
  digital or data matters, such as patient safety, quality of care and
  the licensing of health professionals.
\item
  \textbf{Digital health regulation:} rules made specifically for
  telehealth, health apps, electronic health records, AI diagnostics,
  remote monitoring and other digital health technologies.
\item
  \textbf{International instruments:} references to, alignment with, or
  implementation of international treaties, conventions, guidelines or
  standards, such as those of the WHO, the UN or regional bodies.
\item
  \textbf{Emerging technologies:} governance of artificial intelligence,
  machine learning, blockchain, the Internet of Things and other new
  technologies in health.
\end{itemize}

One element in one area of law is a \textbf{cell}, so each document has
300 cells. Every cell asks the question the human coders were given
(Section 4.3): does this document provide evidence that this element is
addressed in this area of law? Evidence is a provision that
substantively addresses the element, within that area of law, rather
than a neighboring element; naming a problem without committing to act
on it is not evidence. The same provision can be evidence in one area of
law and not in another. Table 5 works through six cells.

For each cell the model gives a verdict, evidence or no evidence. For
evidence it also gives the passages it relies on, quoted word for word
with their page, and a confidence between 0 and 1. A cell the model
reports evidence for is \textbf{flagged}.

\subsubsection{3.2 How a model reads a
document}\label{how-a-model-reads-a-document}

All runs used the tool's default mode, which we call multi-turn Direct
mode. The document goes into an opening message, and each element is
then asked as its own question, one at a time. Earlier questions and
answers are not carried forward, so every question goes out together
with the whole document: a run of 100 questions over the two laws sent
between 2.13 and 2.26 million input tokens.

The instruction on what counts as evidence is one sentence: \emph{``For
EACH priority area, search for evidence of legal or regulatory
provisions that actively address, implement, or mandate this Core
Element.''} The model sees each element's title, not its definition in
the framework, and a short description of each area of law. Nothing in
the prompt says when a provision belongs to an area of law or serves an
element. As Section 8 shows, those two questions decide most errors.

\subsubsection{3.3 Checking the quotes}\label{checking-the-quotes}

Every quote is matched against the text of the page it cites, allowing
for small formatting and optical character recognition differences. A
quote that does not match there is searched for in the whole document;
if it is found elsewhere, the page is corrected. Otherwise it is marked
not found. A finding takes the worst result among its quotes, and a
finding with a quote that cannot be found, or with a confidence below
0.6, is marked for human review. A verified quote shows that a passage
exists. It does not show that the passage answers the question.

\subsubsection{3.4 Combining models}\label{combining-models}

The tool combines runs by vote. A cell is \textbf{evidence by majority}
when at least three of the four models flag it; two of four is a tie,
which goes to review. This is what the tool's Consensus page reports. We
also report two other combinations: \textbf{at least two} models, and
the \textbf{union}, where any one model's flag counts.

\subsubsection{3.5 Votes read as
negatives}\label{votes-read-as-negatives}

Some findings open as a negative (``No evidence of X, though \ldots{}'')
and carry no quote, yet the version of the tool that made these runs
stored them as evidence. We count them as no evidence: 13 of the 295
stored evidence votes on the PIPL and 4 of the 166 on the PDPA. New runs
store them correctly.

\subsection{4. Data and the human
coding}\label{data-and-the-human-coding}

\subsubsection{4.1 Documents and models}\label{documents-and-models}

Table 2 describes the two laws. China's Personal Information Protection
Law, adopted in 2021, was read in an English translation published by
China Briefing (China Briefing 2021). Malaysia's Personal Data
Protection Act 2010 was read in the official reprint as at 1 July 2023
(Commissioner of Law Revision 2023). A reprint as at 15 September 2026
has since superseded it (Commissioner of Law Revision 2026). It
incorporates the Personal Data Protection (Amendment) Act 2024, which
adds, among other changes, a duty to appoint a data protection officer,
notification of data breaches and a right to data portability. The
results here describe the Act before those amendments. A third document,
China's national plan for developing artificial intelligence, issued by
the State Council in 2017 and read in an English translation by the
Foundation for Law and International Affairs (State Council 2017), was
run with three models (28 pages). It is used only to compare how much of
a document models cite and has no human coding.

\begin{longtable}[]{@{}
  >{\raggedright\arraybackslash}p{(\linewidth - 4\tabcolsep) * \real{0.5254}}
  >{\raggedleft\arraybackslash}p{(\linewidth - 4\tabcolsep) * \real{0.2373}}
  >{\raggedleft\arraybackslash}p{(\linewidth - 4\tabcolsep) * \real{0.2373}}@{}}
\caption{\textbf{The two coded laws.}}\tabularnewline
\toprule\noalign{}
\begin{minipage}[b]{\linewidth}\raggedright
\end{minipage} & \begin{minipage}[b]{\linewidth}\raggedleft
China PIPL
\end{minipage} & \begin{minipage}[b]{\linewidth}\raggedleft
Malaysia PDPA
\end{minipage} \\
\midrule\noalign{}
\endfirsthead
\toprule\noalign{}
\begin{minipage}[b]{\linewidth}\raggedright
\end{minipage} & \begin{minipage}[b]{\linewidth}\raggedleft
China PIPL
\end{minipage} & \begin{minipage}[b]{\linewidth}\raggedleft
Malaysia PDPA
\end{minipage} \\
\midrule\noalign{}
\endhead
\bottomrule\noalign{}
\endlastfoot
Pages & 15 & 100 \\
Words & 6,016 & 24,686 \\
Approximate tokens & 9,000 & 35,000 \\
Cells & 300 & 300 \\
Cells holding evidence (estimated, 95\% interval) & 78.7 (58.9 to 112.3)
& 29.3 (23.4 to 53.6) \\
\end{longtable}

Four models assessed both laws in October 2026, with the tool's default
settings (Table 3). They differ in style: deepseek-flash flagged the
most cells on both laws, qwen3.6-35b and gemini-3.5-flash-lite the
fewest.

\begin{longtable}[]{@{}llrr@{}}
\caption{\textbf{The four models and the cells each
flagged.}}\tabularnewline
\toprule\noalign{}
Model & Reached through & Cells flagged, PIPL & Cells flagged, PDPA \\
\midrule\noalign{}
\endfirsthead
\toprule\noalign{}
Model & Reached through & Cells flagged, PIPL & Cells flagged, PDPA \\
\midrule\noalign{}
\endhead
\bottomrule\noalign{}
\endlastfoot
deepseek-flash & DeepSeek's own API & 104 & 64 \\
gpt-6-luna & Azure & 71 & 40 \\
qwen3.6-35b & a local server (OpenWebUI) & 54 & 28 \\
gemini-3.5-flash-lite & Google's API & 53 & 30 \\
\end{longtable}

Each model also made a second run of both laws with identical settings,
which Section 6 uses. Finally, an earlier set of seven models, run on
the PDPA with one call per element over 16 elements (96 cells), shows
how majority votes behave as models are added.

\subsubsection{4.2 Which cells were
checked}\label{which-cells-were-checked}

Checking all 300 cells of a law by hand was not affordable, and a simple
random sample would have spent most of its budget on cells no model
flagged. We sampled 60 cells per law, stratified by how many of the four
models flagged the cell (none, one, two, three or all four), with the
one-model group split further by which model it was. Groups of five or
fewer cells were taken whole. The rest of the budget went to groups in
proportion to the square root of their size, with at least three cells
per group, spread across the areas of law. Table 4 shows the result.

\begin{longtable}[]{@{}
  >{\raggedright\arraybackslash}p{(\linewidth - 12\tabcolsep) * \real{0.2000}}
  >{\raggedleft\arraybackslash}p{(\linewidth - 12\tabcolsep) * \real{0.1286}}
  >{\raggedleft\arraybackslash}p{(\linewidth - 12\tabcolsep) * \real{0.1286}}
  >{\raggedleft\arraybackslash}p{(\linewidth - 12\tabcolsep) * \real{0.1429}}
  >{\raggedleft\arraybackslash}p{(\linewidth - 12\tabcolsep) * \real{0.1286}}
  >{\raggedleft\arraybackslash}p{(\linewidth - 12\tabcolsep) * \real{0.1286}}
  >{\raggedleft\arraybackslash}p{(\linewidth - 12\tabcolsep) * \real{0.1429}}@{}}
\caption{\textbf{Cells by how many models flagged them, how many were
checked, and how many held evidence.}}\tabularnewline
\toprule\noalign{}
\begin{minipage}[b]{\linewidth}\raggedright
Models flagging the cell
\end{minipage} & \begin{minipage}[b]{\linewidth}\raggedleft
PIPL: cells
\end{minipage} & \begin{minipage}[b]{\linewidth}\raggedleft
checked
\end{minipage} & \begin{minipage}[b]{\linewidth}\raggedleft
held evidence
\end{minipage} & \begin{minipage}[b]{\linewidth}\raggedleft
PDPA: cells
\end{minipage} & \begin{minipage}[b]{\linewidth}\raggedleft
checked
\end{minipage} & \begin{minipage}[b]{\linewidth}\raggedleft
held evidence
\end{minipage} \\
\midrule\noalign{}
\endfirsthead
\toprule\noalign{}
\begin{minipage}[b]{\linewidth}\raggedright
Models flagging the cell
\end{minipage} & \begin{minipage}[b]{\linewidth}\raggedleft
PIPL: cells
\end{minipage} & \begin{minipage}[b]{\linewidth}\raggedleft
checked
\end{minipage} & \begin{minipage}[b]{\linewidth}\raggedleft
held evidence
\end{minipage} & \begin{minipage}[b]{\linewidth}\raggedleft
PDPA: cells
\end{minipage} & \begin{minipage}[b]{\linewidth}\raggedleft
checked
\end{minipage} & \begin{minipage}[b]{\linewidth}\raggedleft
held evidence
\end{minipage} \\
\midrule\noalign{}
\endhead
\bottomrule\noalign{}
\endlastfoot
None & 190 & 22 & 2 & 231 & 27 & 0 \\
One & 32 & 14 & 2 & 27 & 12 & 2 \\
Two & 22 & 7 & 2 & 14 & 7 & 1 \\
Three & 18 & 7 & 5 & 5 & 5 & 3 \\
All four & 38 & 10 & 10 & 23 & 9 & 8 \\
\end{longtable}

\subsubsection{4.3 The coding}\label{the-coding}

Two human coders each coded both samples on their own: all 60 cells of
each law, 120 in all. They were chosen for experience reading regulatory
documents, strong analytical reading and concise writing in English, and
the ability to distinguish a genuine match from a plausible but
insufficient connection, which is the judgment every cell calls for.
Throughout this paper, \emph{coders} means these two people; the four
language models are always called \emph{models}. For each cell the
coders saw the element, the area of law with its description, and up to
four candidate passages pooled from the models that had flagged the
cell. The passages were deduplicated, shuffled and unlabeled, so the
coders did not know which model had suggested a passage, how many had,
or what any model had concluded. They were told to treat the passages as
leads only, to search the document themselves when none qualified or
there were none, and to record a verdict (evidence, no evidence, or,
rarely, unclear), the quote and page of any evidence, and optionally a
reason code. Evidence had to address the element substantively, in the
cell's own area of law, and not a neighboring element.

The coders agreed on 54 of the 60 PIPL cells (Krippendorff's alpha 0.79)
and 56 of the 60 PDPA cells (alpha 0.82). Every one of the 10
disagreements turned on one of two questions: does a provision belong to
this area of law, or only touch on it; and does it do what the element
asks, or only help indirectly? The author, who led the study, saw both
coders' verdicts, quotes and notes side by side, settled each question
once, wrote the answer down as a rule, and applied it to every
disagreement:

\begin{itemize}
\tightlist
\item
  A provision counts for an area of law only when what it regulates
  falls within that area's description. Mentioning the area's subject is
  not enough.
\item
  A provision counts for an element only when what it requires,
  prohibits, permits or establishes is what the element calls for. A
  broader rule that includes the element's case counts; overlap, side
  effects or serving a neighboring element do not.
\end{itemize}

The \textbf{gold verdict} is the coders' shared verdict, or the settled
one where they disagreed, so every one of the 120 gold verdicts was made
by people. No model's verdict is part of the gold. The coders found
evidence in 21 of the 60 PIPL cells and 14 of the 60 PDPA cells.

Table 5 shows six of the checked cells and how they were decided. They
return later in the paper: a cell every model got right, the one cell
where every model was wrong, a lone finding that was right, the cell
behind the union's gain on the PIPL, and the two cells of the PIPL that
no model's first run flagged.

\begin{longtable}[]{@{}
  >{\raggedright\arraybackslash}p{(\linewidth - 6\tabcolsep) * \real{0.2947}}
  >{\raggedright\arraybackslash}p{(\linewidth - 6\tabcolsep) * \real{0.1263}}
  >{\raggedright\arraybackslash}p{(\linewidth - 6\tabcolsep) * \real{0.1263}}
  >{\raggedright\arraybackslash}p{(\linewidth - 6\tabcolsep) * \real{0.4526}}@{}}
\caption{\textbf{Six checked cells and how they were
decided.}}\tabularnewline
\toprule\noalign{}
\begin{minipage}[b]{\linewidth}\raggedright
Cell: law, element, area of law
\end{minipage} & \begin{minipage}[b]{\linewidth}\raggedright
Models that flagged it
\end{minipage} & \begin{minipage}[b]{\linewidth}\raggedright
Gold verdict
\end{minipage} & \begin{minipage}[b]{\linewidth}\raggedright
Why
\end{minipage} \\
\midrule\noalign{}
\endfirsthead
\toprule\noalign{}
\begin{minipage}[b]{\linewidth}\raggedright
Cell: law, element, area of law
\end{minipage} & \begin{minipage}[b]{\linewidth}\raggedright
Models that flagged it
\end{minipage} & \begin{minipage}[b]{\linewidth}\raggedright
Gold verdict
\end{minipage} & \begin{minipage}[b]{\linewidth}\raggedright
Why
\end{minipage} \\
\midrule\noalign{}
\endhead
\bottomrule\noalign{}
\endlastfoot
PDPA, 1.1.3 Collect personal or sensitive data only when necessary and
with informed consent, data protection & All four & Evidence & Section 6
bars processing personal data without consent and limits it to what a
lawful purpose needs; section 40 adds explicit consent for sensitive
data, which includes health data. The coders agreed. \\
PDPA, 1.2.5 Ensure data quality, availability, and accessibility,
cybersecurity & All four & No evidence & Section 9's security principle
protects data from loss, misuse and alteration. That is security, not
ensuring data quality; data accuracy is the subject of a separate data
protection provision. The coders split, and the element rule settled
it. \\
PDPA, 2.3.1 Apply health data governance to emerging technologies, data
protection & gpt-6-luna only & Evidence & The Act applies to automated
processing and treats health information as sensitive data: a broader
rule that includes the element's case. The coders split, and the element
rule settled it. \\
PIPL, 2.2.1 Establish data sharing rules and guidelines, cybersecurity &
gpt-6-luna only & Evidence & Article 40 requires a security assessment
by the cyberspace administration before critical information
infrastructure operators and large processors transfer personal data
abroad. The coders agreed. \\
PIPL, 1.1.2 Collect data with a defined purpose, emerging technologies &
None & Evidence & Article 26 limits images and identifying information
from cameras in public places to maintaining public security, unless the
person consents to another use. The coders agreed. \\
PIPL, 2.2.2 Validate informed consent before sharing data, emerging
technologies & None & Evidence & The same Article 26 requires the
person's consent for any use beyond public security. The coders split,
and the two rules settled it. \\
\end{longtable}

\subsubsection{4.4 Estimates over all
cells}\label{estimates-over-all-cells}

Each checked cell stands for its group: it is weighted by the number of
cells in the group divided by the number checked (Cochran 1977). Recall
(of the cells that hold evidence, the share a reader flagged), precision
(of the cells a reader flagged, the share that hold evidence) and F1
(their harmonic mean) are computed from the weighted counts. Ninety-five
percent intervals come from a bootstrap that resamples within groups. A
group checked in full is kept as it is; a group whose checked cells all
carry the same votes draws its share of evidence from a Jeffreys
distribution (Brown et al.~2001), because resampling identical rows
shows no variation at all; any other group is resampled with one draw
fewer than it has rows, which keeps small groups from understating their
variance (Rao and Wu 1988). Differences between two readers are computed
on the same resamples, which makes them much tighter than either
reader's own interval. Before the coding, the estimator was checked in
simulation against known answers: recall came within about 2.5 points of
the truth, and the 95\% intervals covered it 92\% to 98\% of the time.

Two caveats apply throughout. With 60 checked cells per law, intervals
on single readers are wide. And the share of evidence among cells no
model flagged rests on 22 and 27 checked cells; for that share we give a
Wilson interval (Wilson 1927).

\subsection{5. What several models add}\label{what-several-models-add}

\subsubsection{5.1 If misses were
independent}\label{if-misses-were-independent}

Suppose each of four models missed a fifth of the real evidence, and
missed it at random. Together they would miss only the evidence all four
happened to miss: 0.2 to the fourth power, well under 1\%. That is the
case for ``more eyes catch more'', and it rests on the models missing
different things.

The PIPL shows the opposite. Its four models missed between 23\% and
45\% of the evidence each (Table 6). Had they missed it independently,
the four together would have missed 1.2\%, about 1 of the 79 evidence
cells. They missed 22.0\%, about 17 cells (95\% interval 5 to 53), while
the most thorough model alone missed 23.2\%. Of the evidence that model
missed, about 17 of 18 cells were missed by all four. On the PDPA the
test has nothing to work with: the most thorough model missed about 1
evidence cell, and another model found it. Across both laws, about
nine-tenths of the evidence the most thorough model missed was missed by
every model.

\begin{longtable}[]{@{}
  >{\raggedright\arraybackslash}p{(\linewidth - 12\tabcolsep) * \real{0.2840}}
  >{\raggedleft\arraybackslash}p{(\linewidth - 12\tabcolsep) * \real{0.1111}}
  >{\raggedleft\arraybackslash}p{(\linewidth - 12\tabcolsep) * \real{0.1358}}
  >{\raggedleft\arraybackslash}p{(\linewidth - 12\tabcolsep) * \real{0.1111}}
  >{\raggedleft\arraybackslash}p{(\linewidth - 12\tabcolsep) * \real{0.1111}}
  >{\raggedleft\arraybackslash}p{(\linewidth - 12\tabcolsep) * \real{0.1358}}
  >{\raggedleft\arraybackslash}p{(\linewidth - 12\tabcolsep) * \real{0.1111}}@{}}
\caption{\textbf{Recall, precision and F1, estimated over all 300 cells
of each law.} Intervals are in Appendix B.}\tabularnewline
\toprule\noalign{}
\begin{minipage}[b]{\linewidth}\raggedright
Reader
\end{minipage} & \begin{minipage}[b]{\linewidth}\raggedleft
PIPL recall
\end{minipage} & \begin{minipage}[b]{\linewidth}\raggedleft
PIPL precision
\end{minipage} & \begin{minipage}[b]{\linewidth}\raggedleft
PIPL F1
\end{minipage} & \begin{minipage}[b]{\linewidth}\raggedleft
PDPA recall
\end{minipage} & \begin{minipage}[b]{\linewidth}\raggedleft
PDPA precision
\end{minipage} & \begin{minipage}[b]{\linewidth}\raggedleft
PDPA F1
\end{minipage} \\
\midrule\noalign{}
\endfirsthead
\toprule\noalign{}
\begin{minipage}[b]{\linewidth}\raggedright
Reader
\end{minipage} & \begin{minipage}[b]{\linewidth}\raggedleft
PIPL recall
\end{minipage} & \begin{minipage}[b]{\linewidth}\raggedleft
PIPL precision
\end{minipage} & \begin{minipage}[b]{\linewidth}\raggedleft
PIPL F1
\end{minipage} & \begin{minipage}[b]{\linewidth}\raggedleft
PDPA recall
\end{minipage} & \begin{minipage}[b]{\linewidth}\raggedleft
PDPA precision
\end{minipage} & \begin{minipage}[b]{\linewidth}\raggedleft
PDPA F1
\end{minipage} \\
\midrule\noalign{}
\endhead
\bottomrule\noalign{}
\endlastfoot
deepseek-flash & 0.768 & 0.581 & 0.661 & 0.966 & 0.436 & 0.601 \\
gpt-6-luna & 0.666 & 0.749 & 0.705 & 0.902 & 0.735 & 0.810 \\
qwen3.6-35b & 0.654 & 0.942 & 0.772 & 0.731 & 0.766 & 0.748 \\
gemini-3.5-flash-lite & 0.548 & 0.807 & 0.653 & 0.766 & 0.680 & 0.720 \\
Majority (three or four models) & 0.646 & 0.908 & 0.755 & 0.800 & 0.837
& 0.818 \\
At least two models & 0.726 & 0.733 & 0.729 & 0.868 & 0.606 & 0.714 \\
Union (any model) & 0.780 & 0.558 & 0.651 & 1.000 & 0.425 & 0.596 \\
\end{longtable}

So the union of all four models recalled 1.3 points more than
deepseek-flash alone on the PIPL (95\% interval 0.9 to 1.7) and 3.4
points more on the PDPA (1.9 to 4.3). Each gain is a single checked cell
that only gpt-6-luna found: rules for sharing data, under cybersecurity,
on the PIPL, and applying health data governance to emerging
technologies, under data protection, on the PDPA (Table 5).

\subsubsection{5.2 Why the misses overlap}\label{why-the-misses-overlap}

All four models read the same text, receive the same prompt, see the
same element titles without definitions, and were trained on broadly
similar material. Evidence that is easy for one tends to be easy for
all, and evidence that is hard is hard for the same reasons. The shared
miss on the PIPL is a single provision. Article 26 sets purpose and
consent rules for cameras and identification equipment in public places,
which serve two elements under the regulation of emerging technologies:
collecting data for a defined purpose, and validating consent before
data is shared. Every model quoted the page it sits on, between 9 and 49
times, and deepseek-flash quoted Article 26 itself, for a different
element. None connected it to those two elements. The miss was not a
failure to read; it was an inference no model made. A blind spot that
comes from the task is a blind spot for every model given that task.

\subsubsection{5.3 Where the models differ: individual
extras}\label{where-the-models-differ-individual-extras}

The models differ in which question they think a passage answers, not in
which passages they read. Of the cells the union adds beyond the model
that flagged the most, nearly all rest on a passage that model had
itself quoted for another element or area of law: 5 of 6 on the PIPL, 20
of 21 on the AI plan and 5 of 5 on the PDPA. None cites only pages it
never cited.

These individual extras are mostly wrong. Cells only one model flagged
held evidence 4 times in 26 (Table 7). Their quotes are real: 29 of 32
such findings on the PIPL and 27 of 27 on the PDPA passed the quote
check. The passages exist; they answer a different question. A bigger
panel therefore adds findings mainly in the cells where there is nothing
to find.

\subsubsection{5.4 Agreement tells a reviewer where to
look}\label{agreement-tells-a-reviewer-where-to-look}

What the panel does provide is a graded signal of which findings to
trust (Figure 1, Table 7). Cells all four models flagged held evidence
18 times in 19, cells three flagged 8 in 12, two 3 in 14, one 4 in 26,
and none 2 in 49.

\begin{figure}
\centering
\pandocbounded{\includegraphics[keepaspectratio]{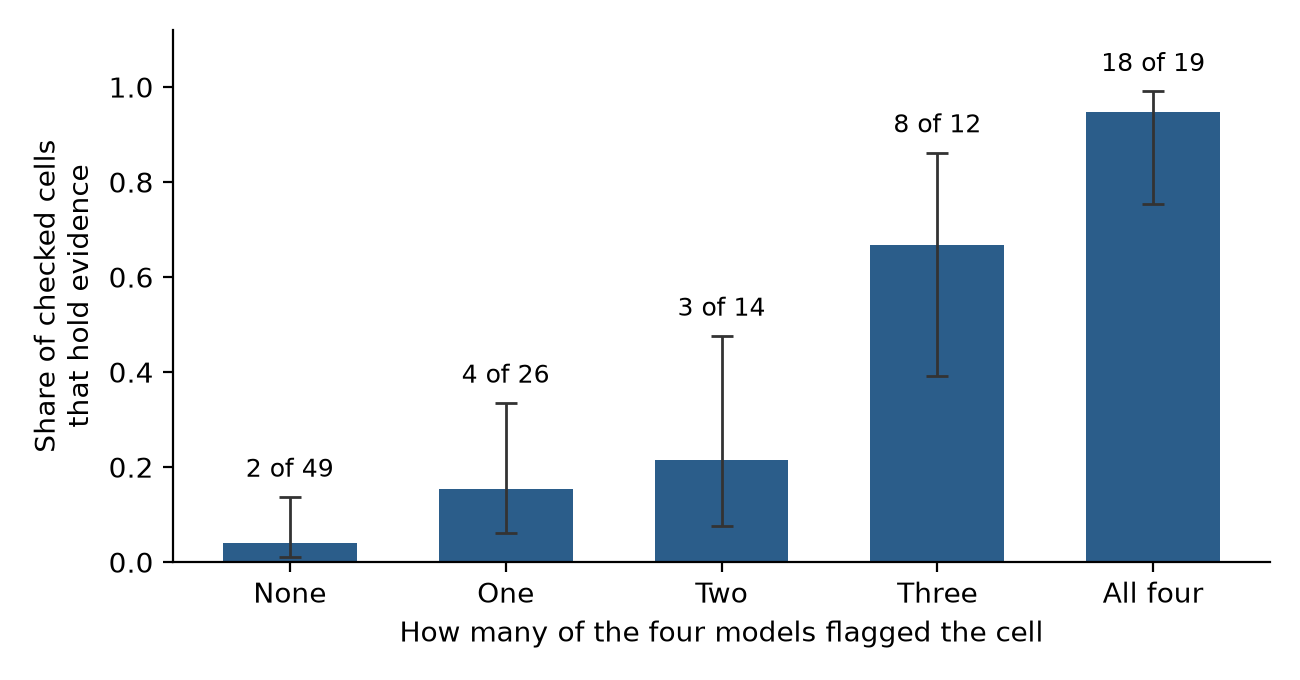}}
\caption{Share of the checked cells that hold evidence, by how many of
the four models flagged the cell, both laws together. Bars show 95\%
Wilson intervals.}
\end{figure}

\begin{longtable}[]{@{}lrrr@{}}
\caption{\textbf{Checked cells holding evidence, by how many models
flagged them.}}\tabularnewline
\toprule\noalign{}
Models flagging the cell & PIPL & PDPA & Both laws \\
\midrule\noalign{}
\endfirsthead
\toprule\noalign{}
Models flagging the cell & PIPL & PDPA & Both laws \\
\midrule\noalign{}
\endhead
\bottomrule\noalign{}
\endlastfoot
All four & 10 of 10 & 8 of 9 & 18 of 19 \\
Three & 5 of 7 & 3 of 5 & 8 of 12 \\
Two & 2 of 7 & 1 of 7 & 3 of 14 \\
One & 2 of 14 & 2 of 12 & 4 of 26 \\
None & 2 of 22 & 0 of 27 & 2 of 49 \\
\end{longtable}

A single model's confidence is a weaker signal. Weighted to all cells,
findings a model rated at least 0.7 confident held evidence about 85\%
of the time, and findings rated 0.4 to 0.69 about 22\%. But agreement
dominates: when one model was alone in a finding, even its findings
rated 0.7 or more held evidence only 1 time in 9.

A council also flags likely fabrications. In extraction from climate
plans, entries more than one model produced were judged genuine far more
often than entries only one model produced (Chong 2026b), and that
pipeline marks an entry as a likely invention when only one model
produced it and its quote cannot be found in the document. In this study
the lone findings mostly quoted real text, so what the council flags
here is misattribution rather than invention. The review it calls for is
the same.

\subsubsection{5.5 Voting can only drop
evidence}\label{voting-can-only-drop-evidence}

A majority vote keeps a cell only when three or four models flag it. It
can drop cells the union holds and can never add one. That makes it a
precision filter: on the PIPL the majority's findings held evidence
90.8\% of the time against 55.8\% for the union, and on the PDPA 83.7\%
against 42.5\%. The price is recall: the majority missed 35.4\% and
20.0\% of the evidence, the union 22.0\% and none. A single precise
model did about as well as the vote. On the PIPL, qwen3.6-35b reached an
F1 of 0.772 against the majority's 0.755; on the PDPA, gpt-6-luna
reached 0.810 against 0.818. The vote is also more stable than any one
model. In the earlier set of seven models, two single models disagreed
on 16.0\% of cells on average, and two three-model majorities with no
model in common on 12.1\%. But the majority did not grow as models were
added (24.3, 23.9 and 23 cells on average for three, five and seven
models), while the union grew from 25.7 cells for one model to 47 for
seven.

\subsubsection{5.6 The base model decides what more models
add}\label{the-base-model-decides-what-more-models-add}

Whether more models raise recall depends on the model they are added to
(Figure 2). Added to deepseek-flash, the most thorough, the other three
added 1.3 and 3.4 points. Added to any of the three more cautious
models, they added between 9.8 and 26.9 points, and every one of those
intervals excludes zero. The companion study of extraction from climate
plans saw the same: there, measured against the council's own consensus
rather than human coding, a precision-dominant model (Grok 4.3,
precision 0.975, recall 0.671) left many items that the consensus
recovered (Chong 2026a). Adding models to a cautious base improves
recall; adding them to an already thorough base does not. Whether the
second half holds for extraction from long adaptation plans has not been
tested against human coders.

\begin{figure}
\centering
\pandocbounded{\includegraphics[keepaspectratio]{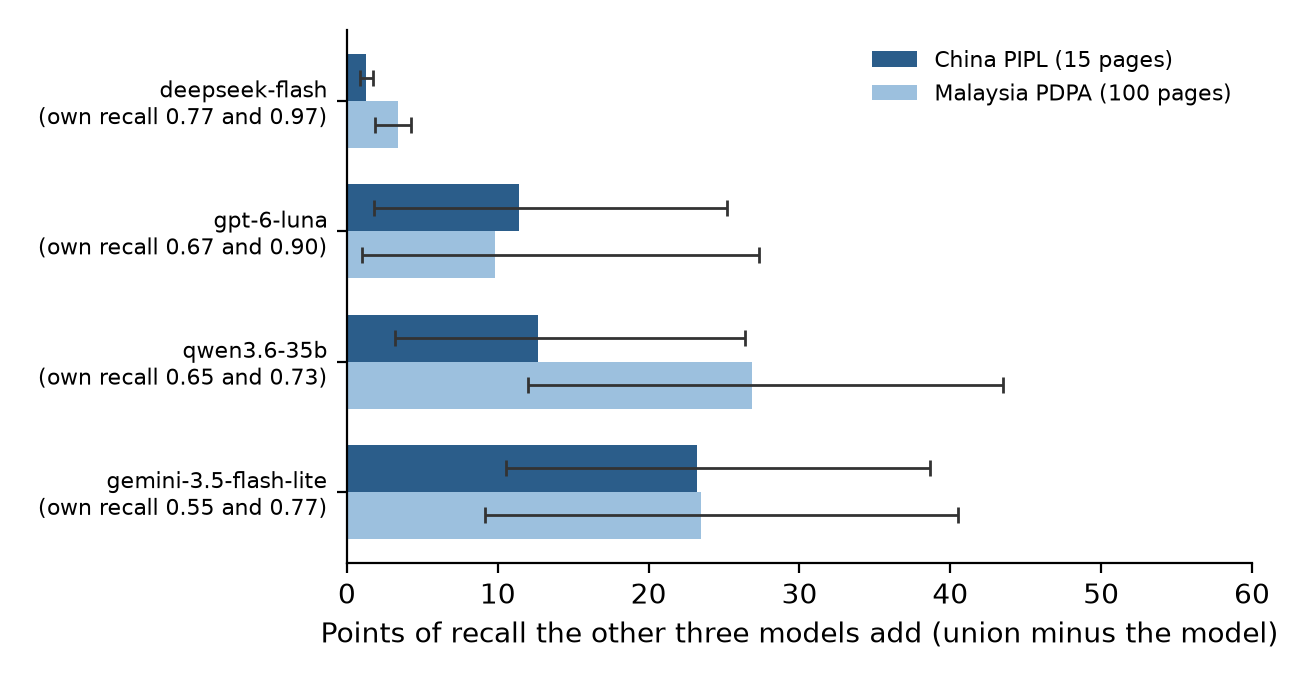}}
\caption{Points of recall the other three models add to each model,
estimated over all cells of each law. Bars show 95\% bootstrap
intervals; each model's own recall on the PIPL and the PDPA is in
parentheses.}
\end{figure}

No model was best on both laws, so the choice of base cannot be made
safely in advance. The best F1 belonged to qwen3.6-35b on the PIPL and
to gpt-6-luna on the PDPA. deepseek-flash had the highest recall on
both, and the lowest precision on both: on the PDPA, more than half of
its findings held no evidence. Running several models insures against
choosing the wrong one.

\subsubsection{5.7 What agreement cannot
fix}\label{what-agreement-cannot-fix}

Agreement is not proof. On the PDPA, all four models cited the Act's
security principle (section 9, which requires steps to protect personal
data from loss, misuse and unauthorized access) as evidence that the law
ensures data quality, availability and accessibility in the
cybersecurity area. The human coders rejected it: keeping data secure is
not ensuring its quality. That was the one unanimous PDPA cell of 9
checked that held no evidence.

Nor can agreement find what every model misses. On the PIPL, the cells
no model flagged held about a fifth of all the evidence. On the PDPA,
none of the 27 checked cells that no model flagged held evidence; the
Wilson interval still allows up to 12.5\% of all 231 such cells to hold
some. Models also agreed with each other far less than the two human
coders did: Krippendorff's alpha across the four models was 0.633 on the
PIPL and 0.639 on the PDPA, against 0.79 and 0.82 for the coders.

\subsection{6. Different models, or different
runs?}\label{different-models-or-different-runs}

\subsubsection{6.1 Two runs of one model disagree
too}\label{two-runs-of-one-model-disagree-too}

Each model's second run used the same document, prompt and settings, so
any difference between its two runs is chance. We measure agreement
between two runs as the share of the cells either run flagged that both
flagged; one minus that share is how much they disagree. Table 8
compares two runs of the same model with runs of two different models
(every pairing of the two models' runs, averaged).

\begin{longtable}[]{@{}lrr@{}}
\caption{\textbf{Of the cells either run flagged, the share both
flagged.}}\tabularnewline
\toprule\noalign{}
Pair of runs & PIPL & PDPA \\
\midrule\noalign{}
\endfirsthead
\toprule\noalign{}
Pair of runs & PIPL & PDPA \\
\midrule\noalign{}
\endhead
\bottomrule\noalign{}
\endlastfoot
Two runs of the same model: average of four models & 0.71 & 0.70 \\
Same model: lowest and highest & 0.64 to 0.76 & 0.62 to 0.80 \\
Two different models: average of six pairs & 0.57 & 0.53 \\
Different models: lowest and highest & 0.47 to 0.65 & 0.37 to 0.64 \\
\end{longtable}

Two runs of one model disagree on about 30\% of the cells either flags;
two different models, on about 45\%. Cohen's kappa over all 300 cells
tells the same story: 0.77 and 0.79 between runs of one model, 0.64 and
0.63 between models. Model diversity is real, but much of the
disagreement between models is already present between two runs of the
same model. Some pairs of different models (qwen3.6-35b and
gemini-3.5-flash-lite, at 0.65 and 0.64) agree as closely as two runs of
one model do.

\subsubsection{6.2 Agreement between two runs is also a
signal}\label{agreement-between-two-runs-is-also-a-signal}

If agreement between models signals reliable findings, agreement between
runs may too. Table 9 estimates, over all cells of both laws, how often
a cell holds evidence when both runs of a model flagged it and when only
one did.

\begin{longtable}[]{@{}
  >{\raggedright\arraybackslash}p{(\linewidth - 6\tabcolsep) * \real{0.3000}}
  >{\raggedleft\arraybackslash}p{(\linewidth - 6\tabcolsep) * \real{0.2000}}
  >{\raggedleft\arraybackslash}p{(\linewidth - 6\tabcolsep) * \real{0.2000}}
  >{\raggedleft\arraybackslash}p{(\linewidth - 6\tabcolsep) * \real{0.3000}}@{}}
\caption{\textbf{How often a flagged cell holds evidence, by whether one
or both runs of a model flagged it.}}\tabularnewline
\toprule\noalign{}
\begin{minipage}[b]{\linewidth}\raggedright
Model
\end{minipage} & \begin{minipage}[b]{\linewidth}\raggedleft
Flagged by both runs
\end{minipage} & \begin{minipage}[b]{\linewidth}\raggedleft
Flagged by one run only
\end{minipage} & \begin{minipage}[b]{\linewidth}\raggedleft
Difference (95\% interval)
\end{minipage} \\
\midrule\noalign{}
\endfirsthead
\toprule\noalign{}
\begin{minipage}[b]{\linewidth}\raggedright
Model
\end{minipage} & \begin{minipage}[b]{\linewidth}\raggedleft
Flagged by both runs
\end{minipage} & \begin{minipage}[b]{\linewidth}\raggedleft
Flagged by one run only
\end{minipage} & \begin{minipage}[b]{\linewidth}\raggedleft
Difference (95\% interval)
\end{minipage} \\
\midrule\noalign{}
\endhead
\bottomrule\noalign{}
\endlastfoot
deepseek-flash & 0.63 & 0.16 & +0.47 (0.16 to 0.68) \\
gpt-6-luna & 0.87 & 0.35 & +0.52 (0.15 to 0.85) \\
qwen3.6-35b & 0.98 & 0.44 & +0.54 (0.28 to 0.78) \\
gemini-3.5-flash-lite & 0.80 & 0.53 & +0.27 (-0.04 to 0.65) \\
\end{longtable}

For three of the four models, cells both runs flagged held evidence far
more often, and the interval excludes zero; for gemini-3.5-flash-lite
the difference points the same way but its interval includes zero. The
sample was drawn on the first runs' votes, so few checked cells separate
the two runs, and these intervals are wide.

\subsubsection{6.3 A second run can recover a shared
miss}\label{a-second-run-can-recover-a-shared-miss}

Pooled over both laws, deepseek-flash's first run recalled 0.822 of the
evidence, and the four models' first runs together 0.840. The two runs
of deepseek-flash together recalled 0.911, and all eight runs 0.920. Two
checked cells make the difference. In its second run, deepseek-flash
connected Article 26 of the PIPL to collecting data for a defined
purpose, which no first run had done, and it also found the one PDPA
cell that only gpt-6-luna had found before. The Article 26 cell stands
for about 9 cells of the PIPL, which is why the gain looks large. No run
of any model connected Article 26 to the second element it serves. A
shared blind spot is not absolute; a different draw sometimes makes the
inference. But two cells are too few to rank a second run against a
second model as a way of buying recall.

\subsubsection{6.4 What this means for a
panel}\label{what-this-means-for-a-panel}

Some of the signal a panel of models provides is available from repeated
runs of one model, at the same cost per run. Different models still
disagree more than runs of one model, so a panel of models also samples
genuinely different readings. Which buys more trust per dollar, more
models or more runs, is open: each model here has only two runs.

\subsection{7. Does document length
matter?}\label{does-document-length-matter}

\subsubsection{7.1 Two questions about a long
document}\label{two-questions-about-a-long-document}

The extraction study's lesson came from a list task. Table 10 sets the
two tasks side by side.

\begin{longtable}[]{@{}
  >{\raggedright\arraybackslash}p{(\linewidth - 4\tabcolsep) * \real{0.2394}}
  >{\raggedright\arraybackslash}p{(\linewidth - 4\tabcolsep) * \real{0.3803}}
  >{\raggedright\arraybackslash}p{(\linewidth - 4\tabcolsep) * \real{0.3803}}@{}}
\caption{\textbf{What each pipeline asks of a long
document.}}\tabularnewline
\toprule\noalign{}
\begin{minipage}[b]{\linewidth}\raggedright
\end{minipage} & \begin{minipage}[b]{\linewidth}\raggedright
Extracting actions from climate plans
\end{minipage} & \begin{minipage}[b]{\linewidth}\raggedright
Checking laws against a framework
\end{minipage} \\
\midrule\noalign{}
\endfirsthead
\toprule\noalign{}
\begin{minipage}[b]{\linewidth}\raggedright
\end{minipage} & \begin{minipage}[b]{\linewidth}\raggedright
Extracting actions from climate plans
\end{minipage} & \begin{minipage}[b]{\linewidth}\raggedright
Checking laws against a framework
\end{minipage} \\
\midrule\noalign{}
\endhead
\bottomrule\noalign{}
\endlastfoot
The question & List every action in this document. & For this element in
this area of law, is there at least one provision? \\
The answer & An open-ended list that grows with the document & One
verdict, with a few supporting quotes \\
When it is complete & When the list covers the whole document; the model
decides when it looks sufficient & When one qualifying passage is
found \\
How length can hurt & The model stops listing early, and parts of the
document go uncovered & Only if the model misses every qualifying
passage, or credits them to another cell \\
What reading in pieces does & Gives each part of the document a complete
pass of its own & Little for the verdict; it multiplies calls and
removes context \\
\end{longtable}

\subsubsection{7.2 Longer documents have more uncited pages, by
topic}\label{longer-documents-have-more-uncited-pages-by-topic}

Coverage does fall with length (Table 11). A page counts as cited when a
model quotes it for a finding.

\begin{longtable}[]{@{}lrrr@{}}
\caption{\textbf{Pages cited.}}\tabularnewline
\toprule\noalign{}
Document & Pages & Cited by one model & Cited by any model \\
\midrule\noalign{}
\endfirsthead
\toprule\noalign{}
Document & Pages & Cited by one model & Cited by any model \\
\midrule\noalign{}
\endhead
\bottomrule\noalign{}
\endlastfoot
China PIPL & 15 & 14 to 15 & 15 \\
China AI development plan & 28 & 9 to 17 & 20 \\
Malaysia PDPA & 100 & 25 to 44 & 52 \\
\end{longtable}

The uncited pages do not suggest that the models simply stopped before
reaching the end. To check this, we divided the PDPA into ten equal
parts of ten pages each. Excluding the opening part, which holds the
cover and the table of contents, each model cited pages in at least
eight of the remaining nine. Together, their citations covered all nine,
including the final part.

What went uncited was mostly introductory, administrative or procedural
material. Of the 48 pages that no model cited:

\begin{itemize}
\tightlist
\item
  12 pages contained the cover, the table of contents and the opening
  provisions on the Act's title and scope.
\item
  29 pages dealt with administration and enforcement, including the
  Commissioner's staffing, the Fund, the Appeal Tribunal, investigations
  and prosecution.
\item
  7 pages covered procedures within the main rules, such as registration
  certificates and notices refusing access.
\end{itemize}

The AI plan showed a similar pattern. Its eight uncited pages covered
introductory material, research topics and technologies, consumer
devices, and AI in public administration. Seven of those pages did not
mention health, medical care, privacy, personal data or ethics.

This pattern suggests that what a page contained mattered more than
where it appeared in the document. The models continued citing material
near the end, while leaving other pages uncited throughout. That does
not prove they found every relevant provision, but it does argue against
a simple explanation that they stopped early.

An uncited page therefore means different things in different tasks.
When the task is to list every action in a plan, an uncited chapter
containing actions is a warning that something may have been missed.
When the task is to find legal provisions that meet specific governance
requirements, some pages may contain nothing relevant to those
requirements. Fewer cited pages do not, by themselves, show that less
relevant evidence was found.

\subsubsection{7.3 Recall, precision and wrong findings by
law}\label{recall-precision-and-wrong-findings-by-law}

Table 12 compares each model's two standard runs, averaged, on the short
and the long law. Besides recall and precision it gives the
\textbf{wrong-finding rate}: of the cells that hold no evidence, the
share a model flagged. Unlike precision, this rate does not depend on
how much evidence a document holds.

\begin{longtable}[]{@{}
  >{\raggedright\arraybackslash}p{(\linewidth - 8\tabcolsep) * \real{0.1750}}
  >{\raggedright\arraybackslash}p{(\linewidth - 8\tabcolsep) * \real{0.2625}}
  >{\raggedleft\arraybackslash}p{(\linewidth - 8\tabcolsep) * \real{0.1250}}
  >{\raggedleft\arraybackslash}p{(\linewidth - 8\tabcolsep) * \real{0.1250}}
  >{\raggedleft\arraybackslash}p{(\linewidth - 8\tabcolsep) * \real{0.3125}}@{}}
\caption{\textbf{The short law against the long law (average of each
model's two standard runs).} The last column gives the long law minus
the short law, with its 95\% interval.}\tabularnewline
\toprule\noalign{}
\begin{minipage}[b]{\linewidth}\raggedright
Measure
\end{minipage} & \begin{minipage}[b]{\linewidth}\raggedright
Model
\end{minipage} & \begin{minipage}[b]{\linewidth}\raggedleft
PIPL (15 pages)
\end{minipage} & \begin{minipage}[b]{\linewidth}\raggedleft
PDPA (100 pages)
\end{minipage} & \begin{minipage}[b]{\linewidth}\raggedleft
Long minus short
\end{minipage} \\
\midrule\noalign{}
\endfirsthead
\toprule\noalign{}
\begin{minipage}[b]{\linewidth}\raggedright
Measure
\end{minipage} & \begin{minipage}[b]{\linewidth}\raggedright
Model
\end{minipage} & \begin{minipage}[b]{\linewidth}\raggedleft
PIPL (15 pages)
\end{minipage} & \begin{minipage}[b]{\linewidth}\raggedleft
PDPA (100 pages)
\end{minipage} & \begin{minipage}[b]{\linewidth}\raggedleft
Long minus short
\end{minipage} \\
\midrule\noalign{}
\endhead
\bottomrule\noalign{}
\endlastfoot
Recall & deepseek-flash & 0.82 & 0.98 & +0.16 (-0.32 to +0.30) \\
& gpt-6-luna & 0.72 & 0.90 & +0.18 (-0.26 to +0.37) \\
& gemini-3.5-flash-lite & 0.53 & 0.78 & +0.25 (-0.13 to +0.46) \\
& qwen3.6-35b & 0.59 & 0.69 & +0.09 (-0.25 to +0.32) \\
Precision & deepseek-flash & 0.60 & 0.47 & -0.12 (-0.31 to +0.08) \\
& gpt-6-luna & 0.81 & 0.78 & -0.03 (-0.25 to +0.15) \\
& gemini-3.5-flash-lite & 0.81 & 0.68 & -0.13 (-0.34 to +0.10) \\
& qwen3.6-35b & 0.88 & 0.75 & -0.13 (-0.32 to +0.05) \\
Wrong-finding rate & deepseek-flash & 19.7\% & 11.9\% & -7.7 points
(-16.1 to -0.8) \\
& gpt-6-luna & 5.9\% & 2.8\% & -3.1 points (-8.3 to +0.9) \\
& gemini-3.5-flash-lite & 4.4\% & 3.9\% & -0.5 points (-5.4 to +3.8) \\
& qwen3.6-35b & 2.8\% & 2.5\% & -0.4 points (-3.5 to +2.5) \\
\end{longtable}

Recall was higher on the long law for every model, but the data cannot
show that it held: every interval spans zero widely, and a fall of up to
about 0.3 cannot be ruled out. The long law's recall also rests on the
27 checked cells no model flagged, none of which held evidence.
Precision fell for every model, and the lower share of evidence explains
it: the coders' estimate is 29 evidence cells of 300 on the long law
against 79 on the short one, and when evidence is rarer the same rate of
mistakes is a larger share of what a model reports. That rate did not
rise. It is lower on the long law for every model, its interval excludes
zero for deepseek-flash, and no interval allows a rise of more than 3.8
points.

\subsubsection{7.4 The misses were on pages the models had
read}\label{the-misses-were-on-pages-the-models-had-read}

On the PDPA, the models missed evidence in checked cells 11 times, and
each time the model had cited the page holding that evidence for some
other cell. The provision all four models missed on the PIPL sits on a
page every model quoted (Section 5.2). Reading in smaller pieces does
not change what a model decides a passage is evidence of.

\subsubsection{7.5 Where the extraction intuition does apply: the quotes
inside a
finding}\label{where-the-extraction-intuition-does-apply-the-quotes-inside-a-finding}

A model gives about as many quotes per finding on the short law as on
the long one (Table 13). That is what stopping once the answer looks
sufficient would look like, at the scale of one cell. It could also mean
that a cell simply has about as many relevant provisions in either law;
these data cannot tell the two apart. For the question ``is there
evidence?'' either reading is harmless. For the question ``where is all
of it?'', the first means the quotes are incomplete. The gold verdicts
score cells, not quotes, so this is unmeasured. If people will read the
quotes as the inventory of evidence, this is where reading in pieces
could help.

\begin{longtable}[]{@{}lrr@{}}
\caption{\textbf{Quotes per finding, average of each model's two
standard runs.}}\tabularnewline
\toprule\noalign{}
Model & PIPL & PDPA \\
\midrule\noalign{}
\endfirsthead
\toprule\noalign{}
Model & PIPL & PDPA \\
\midrule\noalign{}
\endhead
\bottomrule\noalign{}
\endlastfoot
deepseek-flash & 3.62 & 3.67 \\
gpt-6-luna & 2.26 & 2.62 \\
qwen3.6-35b & 2.05 & 2.36 \\
gemini-3.5-flash-lite & 1.42 & 1.63 \\
\end{longtable}

\subsubsection{7.6 What reading in pieces would
cost}\label{what-reading-in-pieces-would-cost}

Reading in pieces would multiply calls: one question per element per
piece, about four times as many with 30-page pieces on the PDPA. It
would also raise the risk of wrong findings. The verdict is ``at least
one provision'', so a cell would be flagged if any piece flags it, and
every piece is another chance to report evidence where there is none;
this is reasoning from the verdict rule, not something we measured. And
each piece would be judged without the rest of the law, whose
definitions and exemptions apply throughout. One case needs pieces
regardless: a document longer than the model's context window, which the
tool today refuses rather than splits.

\subsection{8. The main error is attribution: a test of written
rules}\label{the-main-error-is-attribution-a-test-of-written-rules}

When a coder found no evidence in a checked cell and gave a reason, it
was almost always attribution: the candidate passages served another
element (53 times across both laws and both coders) or belonged to
another area of law (49), against 4 other reasons. The models' errors
look the same. The lone findings of Section 5.3 quote real passages
filed under the wrong cell, and the unanimous error of Section 5.7
credits a security rule to data quality.

The prompt says nothing about either question (Section 3.2), so we
tested adding the two written rules of Section 4.3 to every question, in
an opt-in ``strict'' mode, together with an instruction to report a
close miss as no evidence while naming it. Each model ran once with the
rules (deepseek-flash twice), and each strict run was compared with both
of the model's standard runs on the same checked cells (Table 14).

\begin{longtable}[]{@{}llrr@{}}
\caption{\textbf{Pooled over both laws: each model's two standard runs
(first run and rerun) against its strict runs.}}\tabularnewline
\toprule\noalign{}
Model & Measure & Standard runs & Strict runs \\
\midrule\noalign{}
\endfirsthead
\toprule\noalign{}
Model & Measure & Standard runs & Strict runs \\
\midrule\noalign{}
\endhead
\bottomrule\noalign{}
\endlastfoot
deepseek-flash & Precision & 0.52 and 0.58 & 0.73 and 0.83 \\
& Recall & 0.82 and 0.91 & 0.83 and 0.75 \\
& F1 & 0.64 and 0.71 & 0.78 and 0.79 \\
gpt-6-luna & Precision & 0.74 and 0.86 & 0.84 \\
& Recall & 0.73 and 0.81 & 0.69 \\
& F1 & 0.74 and 0.84 & 0.75 \\
qwen3.6-35b & Precision & 0.88 and 0.79 & 0.90 \\
& Recall & 0.67 and 0.56 & 0.66 \\
& F1 & 0.76 and 0.66 & 0.76 \\
gemini-3.5-flash-lite & Precision & 0.76 and 0.76 & 0.80 \\
& Recall & 0.61 and 0.59 & 0.61 \\
& F1 & 0.67 and 0.67 & 0.69 \\
\end{longtable}

The rules helped the model that reported too much. deepseek-flash's
precision rose in both strict runs, and the gain holds against both of
its standard runs. Its F1 gain does not hold against its rerun, because
the rules also cost it some recall. For the three models that were
already precise, no change in precision or F1 could be told from zero
against their first run. Against their reruns, gpt-6-luna looks worse
and qwen3.6-35b better, but those reruns had themselves scored 0.10 F1
above and 0.11 below the first runs with no change of settings, so the
gaps are the size of chance. The one change that held against both
standard runs was a loss of recall for gpt-6-luna.

The rules also changed where findings were filed, which is what they are
about. Across both laws, deepseek-flash's findings under general health
law fell from 9 and 8 in its standard runs to none in either strict run,
and those under international instruments from 16 and 19 to 7 and 7.
gemini-3.5-flash-lite's findings moved out of data protection, from 58
and 62 cells to 49, and into cybersecurity, from 5 and 6 to 17. That
fits the rule that the security duties in a data protection law are
cybersecurity provisions, though one strict run cannot show that the
rule caused it. Among the 120 checked cells the trade shows cell by
cell. deepseek-flash's two strict runs left out 11 and 13 wrong findings
that both of its standard runs had made, and dropped 2 and 5 real
findings that at least one standard run had. For the precise models the
two sides came out about even: gpt-6-luna left out 4 wrong findings and
dropped 5 real ones, gemini-3.5-flash-lite 3 and 6 (while finding 3 real
ones its standard runs had missed), and qwen3.6-35b 1 and 3. Five
checked cells were dropped by two models each, which makes them the
first place to look when rewording the rules.

Writing the rules down is worth doing for the people who apply them;
handing them to the models is an option to test per model, not a cure.

\subsection{9. Discussion: a workflow for models and
people}\label{discussion-a-workflow-for-models-and-people}

The study supports using several models as a cross-check, not as a
substitute for judgment. Their greatest value is helping people decide
what to trust provisionally, what to question, and where to look more
closely. Agreement maps onto those three directly: cells every model
flags can be trusted provisionally, with spot checks; cells where the
models disagree are the ones to question; and cells no model flags are
where to look more closely, because that is where shared blind spots
hide. In practice that suggests the following division of labor.

\begin{enumerate}
\def\labelenumi{\arabic{enumi}.}
\tightlist
\item
  \textbf{Run three or four models with different styles} over the
  complete document, including at least one thorough model, and check
  that every run covered every element. A thorough model sets the
  ceiling on recall; the others provide the agreement signal and insure
  against a poor choice of base.
\item
  \textbf{Let agreement sort the cells.} Accept cells all models flag,
  with spot checks. Send every cell where they disagree to a person:
  here, 72 of the PIPL's 300 cells (24\%) and 46 of the PDPA's (15\%).
  Accept cells no model flags, but audit a random sample, because that
  is where shared blind spots hide; on the PIPL it held about a fifth of
  the evidence. At an assumed 2 to 3 minutes per disputed cell, which we
  did not measure, the review takes about 1.5 to 2.3 hours for the PDPA
  and 2.4 to 3.6 hours for the PIPL.
\item
  \textbf{Write down the attribution rules} and apply them the same way
  across documents and reviewers. They settled every disagreement
  between our coders.
\item
  \textbf{Record decisions where the tool can reuse them,} so they carry
  over to exports and later runs.
\item
  \textbf{Check the inputs.} In one build of the PDPA comparison, an
  earlier test run was selected by mistake: its model name differed from
  the intended one only by ``-lite'', and it covered 16 of the 50
  elements. The comparison page gave no warning then; it now flags runs
  that cover different elements. Only a person checking what went in
  would have caught it.
\end{enumerate}

On reading in pieces: for verdicts of the form ``at least one
provision'', the evidence here gives no reason to do it and some reasons
not to. It becomes necessary for documents longer than the context
window, and it may help if the quotes are meant as a complete inventory.
Both cases need testing against human coding before a tool relies on
them, with attention to the rate of wrong findings, not only recall.

\subsection{10. Limitations}\label{limitations}

\begin{itemize}
\tightlist
\item
  \textbf{Two laws.} Both are data protection statutes read in English;
  China's law was read in translation, while its Chinese text is the
  authoritative one. The Malaysian results describe the Act before its
  2024 amendments, so cells on breach notification, data protection
  officers or data portability could be decided differently on the
  current text. A long policy plan, whose chapters are mostly
  substantive, could behave differently, and length is mixed up here
  with the kind of law and how much evidence it holds. Nothing in this
  study isolates length.
\item
  \textbf{Nothing near the context window.} The longer law is about
  35,000 tokens, far inside every model's window.
\item
  \textbf{Few checked cells.} Sixty per law, so single figures have wide
  intervals, and a few heavily weighted cells move them. The shared-miss
  estimate on the PIPL rests on two checked cells holding the same
  provision.
\item
  \textbf{The gold depends on two rulings.} Ten of the 120 checked cells
  were settled by the author's two rules, and the strict-rules test uses
  the same rules. The author made these rulings while also running the
  study. The other 110 were agreed by the two human coders working
  independently. Scoring the strict runs again without the 10 cells the
  author ruled on, or with either coder's own verdicts in their place,
  moved each model's average change from the strict rules (in precision,
  recall and F1) by about 0.05 at most and reversed none, but the rules
  were written from these two laws, so they still need testing on a new,
  independently coded document.
\item
  \textbf{Coders saw candidate passages.} Unlabeled passages from the
  models were shown as leads. They may draw a coder toward what the
  models found; the coders were told to search further, and cells no
  model flagged came with no candidates at all.
\item
  \textbf{Two runs per model.} The comparison of runs with models rests
  on one rerun each.
\item
  \textbf{One reading mode.} All runs sent the whole document with every
  question. Retrieval-based reading, which the tool also offers, chooses
  what to read and was not tested.
\item
  \textbf{Model names are not versions.} Providers change models without
  notice; the figures describe the models as they answered in October
  2026.
\end{itemize}

\subsection{11. Conclusion}\label{conclusion}

Several models did not find much evidence that a thorough model missed,
because their misses were shared: they read the same text, with the same
prompt, and failed at the same inferences. Their extra findings were
individual and mostly wrong. What a panel buys is a graded signal of
which findings to trust, insurance against choosing a poor model, and a
flag on lone findings for a person to check; the vote that reports its
consensus trades recall for precision. Long documents did not need to be
read in pieces for these verdicts, because one qualifying provision
settles a cell and the pages left uncited held little the framework asks
about. The errors that matter are attributions, which neither more
models, more runs nor smaller pieces remove, and which a person writing
down the rules can. Used this way, several models are a cross-check on
people's judgment, not a substitute for it: they show what to trust
provisionally, what to question, and where to look more closely.

\subsection{Acknowledgments}\label{acknowledgments}

We thank the two coders for their careful, independent coding of the 120
cells on which every estimate in this paper rests. The United Nations
University Campus Computing Centre paid the cost of the language models'
API use.

\subsection{Appendix A. Terms}\label{appendix-a.-terms}

{\def\LTcaptype{none} 
\begin{longtable}[]{@{}
  >{\raggedright\arraybackslash}p{(\linewidth - 2\tabcolsep) * \real{0.3286}}
  >{\raggedright\arraybackslash}p{(\linewidth - 2\tabcolsep) * \real{0.6714}}@{}}
\toprule\noalign{}
\begin{minipage}[b]{\linewidth}\raggedright
Term
\end{minipage} & \begin{minipage}[b]{\linewidth}\raggedright
Meaning
\end{minipage} \\
\midrule\noalign{}
\endhead
\bottomrule\noalign{}
\endlastfoot
Cell & One core element of the framework in one area of law; 300 per
document. \\
Coder & One of the two people who checked the sampled cells. Models are
never called coders. \\
Gold verdict & For a checked cell, the coders' shared verdict, or the
author's ruling where they disagreed. \\
Flag & A model reports evidence for a cell. \\
Evidence & A provision that does what the element asks, within that area
of law, as the human coders judged it. \\
Recall & Of the cells that hold evidence, the share a reader flagged. \\
Precision & Of the cells a reader flagged, the share that hold
evidence. \\
F1 & The harmonic mean of recall and precision. \\
Union & A cell counts as flagged if any model flags it. \\
Majority & A cell counts as flagged if at least three of the four models
flag it. \\
Wrong-finding rate & Of the cells that hold no evidence, the share a
model flagged. \\
Attribution & Deciding which element and which area of law a passage
serves. \\
Standard run & A run with the default prompt; each model has a first run
and a rerun. \\
\end{longtable}
}

\subsection{Appendix B. Intervals for Table
6}\label{appendix-b.-intervals-for-table-6}

Ninety-five percent bootstrap intervals, estimated over all 300 cells of
each law.

{\def\LTcaptype{none} 
\begin{longtable}[]{@{}
  >{\raggedright\arraybackslash}p{(\linewidth - 8\tabcolsep) * \real{0.3067}}
  >{\raggedleft\arraybackslash}p{(\linewidth - 8\tabcolsep) * \real{0.1733}}
  >{\raggedleft\arraybackslash}p{(\linewidth - 8\tabcolsep) * \real{0.1733}}
  >{\raggedleft\arraybackslash}p{(\linewidth - 8\tabcolsep) * \real{0.1733}}
  >{\raggedleft\arraybackslash}p{(\linewidth - 8\tabcolsep) * \real{0.1733}}@{}}
\toprule\noalign{}
\begin{minipage}[b]{\linewidth}\raggedright
Reader
\end{minipage} & \begin{minipage}[b]{\linewidth}\raggedleft
PIPL recall
\end{minipage} & \begin{minipage}[b]{\linewidth}\raggedleft
PIPL precision
\end{minipage} & \begin{minipage}[b]{\linewidth}\raggedleft
PDPA recall
\end{minipage} & \begin{minipage}[b]{\linewidth}\raggedleft
PDPA precision
\end{minipage} \\
\midrule\noalign{}
\endhead
\bottomrule\noalign{}
\endlastfoot
deepseek-flash & 0.54 to 0.93 & 0.45 to 0.70 & 0.53 to 0.97 & 0.33 to
0.57 \\
gpt-6-luna & 0.44 to 0.84 & 0.57 to 0.87 & 0.47 to 0.96 & 0.52 to
0.87 \\
qwen3.6-35b & 0.43 to 0.82 & 0.76 to 1.00 & 0.37 to 0.82 & 0.52 to
0.90 \\
gemini-3.5-flash-lite & 0.36 to 0.70 & 0.60 to 0.94 & 0.39 to 0.86 &
0.46 to 0.83 \\
Majority & 0.43 to 0.81 & 0.71 to 1.00 & 0.41 to 0.90 & 0.60 to 0.92 \\
At least two & 0.50 to 0.88 & 0.56 to 0.86 & 0.45 to 0.93 & 0.43 to
0.72 \\
Union & 0.54 to 0.94 & 0.44 to 0.67 & 0.55 to 1.00 & 0.32 to 0.55 \\
\end{longtable}
}

\subsection{References}\label{references}

Aggarwal, Vikram; Kulkarni, Jay; Mascarenhas, Aditi; Narang, Aakriti;
Raman, Siddarth; Shah, Ajay; Thomas, Susan (2025). Information
Extraction From Fiscal Documents Using LLMs. \emph{arXiv:2511.10659; AI
for Financial Inclusion workshop, ACM ICAIF 2025}.

Artstein, R.; Poesio, M. (2008). Inter-coder agreement for computational
linguistics. \emph{Computational Linguistics}, 34(4), 555-596.

Brown, Lawrence D.; Cai, T. Tony; DasGupta, Anirban (2001). Interval
Estimation for a Binomial Proportion. \emph{Statistical Science}, 16(2),
101-133.

China Briefing (2021). Personal Information Protection Law of the
People's Republic of China. \emph{English translation of the law adopted
on August 20, 2021, in force from November 1, 2021}.
\url{https://www.china-briefing.com/news/wp-content/uploads/2021/08/Personal-Information-Protection-Law-of-the-Peoples-Republic-of-China.pdf}

Chong, Ng S. T. (2026a). Cost-Quality Frontiers in LLM-Assisted Policy
Analysis: Multi-Model Consensus for Structured Extraction and Evaluative
Scoring. \emph{SSRN working paper, abstract 6970018}.
\url{https://papers.ssrn.com/sol3/papers.cfm?abstract_id=6970018}

Chong, Ng S. T. (2026b). Stopping Short: Detecting Omissions, Improving
Coverage, and Calibrating Trust in LLM Extraction from Climate Policy
Documents. \emph{SSRN working paper, abstract 7457638}.
\url{https://papers.ssrn.com/sol3/papers.cfm?abstract_id=7457638}

Cochran, William G. (1977). Sampling Techniques. \emph{3rd edition. New
York: John Wiley and Sons}.

Commissioner of Law Revision, Malaysia (2023). Laws of Malaysia, Act
709: Personal Data Protection Act 2010. \emph{Reprint as at 1 July 2023,
the version this study read}.
\url{https://www.investmalaysia.gov.my/media/3x4fsqum/personal-data-protection-act-2010.pdf}

Commissioner of Law Revision, Malaysia (2026). Laws of Malaysia, Act
709: Personal Data Protection Act 2010. \emph{Reprint as at 15 September
2026, incorporating the Personal Data Protection (Amendment) Act 2024,
Act A1727}.
\url{https://www.pdp.gov.my/ppdpv1/en/akta/personal-data-protection-act-2010-reprint/}

Gilardi, F.; Alizadeh, M.; Kubli, M. (2023). ChatGPT outperforms
crowd-workers for text-annotation tasks. \emph{Proceedings of the
National Academy of Sciences}, 120(30), e2305016120.

Kamen, Ariel; Kamen, Yakov (2025). Majority Rules: LLM Ensemble is a
Winning Approach for Content Categorization. \emph{arXiv:2511.15714}.

Kamradt, Gregory (2023). Needle in a Haystack: Pressure Testing LLMs.
\url{https://github.com/gkamradt/LLMTest_NeedleInAHaystack}.

Krippendorff, K. (2004). Content Analysis: An Introduction to Its
Methodology.

Liu, Nelson F.; Lin, Kevin; Hewitt, John; Paranjape, Ashwin; Bevilacqua,
Michele; Petroni, Fabio; Liang, Percy (2024). Lost in the Middle: How
Language Models Use Long Contexts. \emph{Transactions of the Association
for Computational Linguistics}, 12, 157-173.

Rao, J. N. K.; Wu, C. F. J. (1988). Resampling Inference with Complex
Survey Data. \emph{Journal of the American Statistical Association},
83(401), 231-241.

Schroeder, Hope; Roy, Deb; Kabbara, Jad (2025). Just Put a Human in the
Loop? Investigating LLM-Assisted Annotation for Subjective Tasks.
\emph{Findings of the Association for Computational Linguistics: ACL
2025}. \url{https://aclanthology.org/2025.findings-acl.1323/}

State Council of the People's Republic of China (2017). A New Generation
of Artificial Intelligence Development Plan. \emph{State Council
Document {[}2017{]} No.~35, issued July 8, 2017. English translation by
the Foundation for Law and International Affairs}.
\url{https://www.unodc.org/res/ji/import/policy_papers/china_ai_strategy/china_ai_strategy.pdf}

Transform Health (n.d.). Health Data Governance Principles, and the
National Legislative and Regulatory Assessment Tool.
\url{https://healthdatagovernance.org/assessment-tool/}, accessed
October 18, 2026.

Verga, Pat; Hofstätter, Sebastian; Althammer, Sophia; Su, Yixuan;
Piktus, Aleksandra; Arkhangorodsky, Arkady; Xu, Minjie; White, Naomi;
Lewis, Patrick (2024). Replacing Judges with Juries: Evaluating LLM
Generations with a Panel of Diverse Models. \emph{arXiv:2404.18796}.

Wang, X.; Wei, J.; Schuurmans, D.; others (2023). Self-consistency
improves chain of thought reasoning in language models.
\emph{Proceedings of ICLR 2023}.

Wilson, Edwin B. (1927). Probable Inference, the Law of Succession, and
Statistical Inference. \emph{Journal of the American Statistical
Association}, 22(158), 209-212.

Xu, D.; Chen, W.; Peng, W.; others (2024). Large language models for
generative information extraction: a survey. \emph{Frontiers of
Information Technology \& Electronic Engineering}, 25(11), 1471-1498.

Yang, Chen; Zhang, Xianyang; Chen, Jun (2025). Large Language Model
Consensus Substantially Improves the Cell Type Annotation Accuracy for
scRNA-seq Data. \emph{bioRxiv 2025.04.10.647852}.

Zhang, Congjing; Bao, Ruoxuan; Li, Jingyu; Ackerman, Yoav; Huang, Shuai;
Su, Yanfang (2026). A Role-Based LLM Framework for Structured
Information Extraction from Healthy Food Policies.
\emph{arXiv:2604.01529}.

\end{document}